\documentstyle[epsfbox,epsfig,macros,theapa,11pt,twoside]{article}
\def\psscale{0.45}
\let\prg=\sf

\newcommand{\auth}[1]{{\bf#1}}

\renewcommand{\epsfsize}[2]{\psscale#1}
\newcommand{\epscap}[2]{\hbox{ }\hfill\hbox{\hss\lower4ex\hbox{(#2)}
\vtop{\hbox{\ }\epsfbox{#1.eps}}}\hfill\hbox{ }}
\begin{document}
\title{Parsing and Generation\\with Tabulation and Compilation\thanks{%
1st Workshop on `Tabulation in Parsing and Deduction' (TAPD\,'98),
pp.\,26-35.}}
\author{HASIDA K\^oiti\\
Electrotechnical Laboratory\\
1-1-4 Umezono, Tukuba,\\Ibaraki 305, Japan.\\
{\sl hasida@etl.go.jp}
\and
MIYATA Takashi\\
Nara Institute of Science and Technology\\
8916-5, Takayama, Ikoma\\Nara 630-01, Japan\\
{\sl takashi@is.aist-nara.ac.jp}}
\date{}
\maketitle

\section{Introduction}

The standard tabulation techniques for logic programming
presuppose fixed order of computation:
top-down and left-to-right in OLDT \cite{tamaki&sato86}
and bottom-up and left-to-right in magic set \cite{beeri&ramakrishnan87}.
This makes it hard to deal with problems in which
the distribution of input information is diverse and
accordingly diverse flow of information is required.
Consider natural language understanding for instance.
In one case, syntactic information may be missing due to unknown words
but semantic information may be abundant thanks to extralinguistic contexts,
whereas in another case the situation may be the opposite.
These cases should be treated by drastically different
processing orders for the sake of efficiency.
Integrated treatment of natural language
understanding and production also calls for
flexible control sensitive to various computational contexts.

Some data-driven control should be introduced in order to deal
with such diverse contexts.
A concern of this paper is hence tabulation in data-driven
transformation of constraints.
Another concern is optimization of such computation.
Standard optimization methods in natural language processing
and logic programming include accessibility check and last-call optimization.
The former is employed in left-corner parsing
and semantic-head-driven generation \shortcite{shieber&90}, among others.
The latter is not only used in Earley parsing and left-corner parsing,
but also is a major factor in WAM \cite{warren83,aitkaci91}.
In what follows we will address a data-driven method
of constraint transformation with a sort of compilation
which subsumes accessibility check and last-call optimization.

\section{Dependency Reduction}

A Horn-clause program is regarded as a forest of {\bf program trees}.
A program tree is a candidate for a proof tree.
Namely, in a program tree,
each node is (an instances of) a clause, the root node is a top clause
(the program may have several top clauses),
and each negative (body) literal is linked with a positive (head)
literal carrying the same predicate.
A program tree is a proof tree iff
it contains no contradiction among unified terms.
For instance, \fig{pqprogtree}%
\epsfig{pqprogtree}{A program tree and dependency paths.}
is a program tree, but not a proof tree, of the following program.
\begin{itemz}
\item[(A)] {\prg \LA\ p(X) \AND\ q(U,X,Y) \AND\ r(Y,Z).}
\item[(B)] {\prg p(X) \LA\ X=a.}
\item[(C)] {\prg p(X) \LA\ X=f(Y) \AND\ p(Y).}
\item[(D)] {\prg q(U,X,X) \LA\ U=b.}
\item[(E)] {\prg q(U,X,Z) \LA\ U=g(V) \AND\ q(V,X,Y) \AND\ Y=f(Z).}
\end{itemz}
The definition clauses of {\prg r} are omitted for simplicity.

The thick dim curves in \fig{pqprogtree} are {\bf dependency paths}.
A dependency path is a sequence of variables unified with each other.
A program tree is a proof tree if it has no dependency path
connecting two non-variable terms.
The computational procedure we propose here, {\bf dependency reduction},
is to transform the program by eliminating
dependency paths connecting two non-variable terms,
while preserving the proof trees.

A dependency path connecting two non-variable terms in a clause
can be eliminated in two ways.
First, if the two terms are unifiable, they are unified and eliminated;
this is a valid operation if each variable appears at most twice in a clause,
which we assume without loss of generality.
Second, when the two terms are not unifiable,
the clause is deleted without deleting any proof trees.

We must shorten dependency paths before eliminating them.
A dependency path can be shortened by moving a predicate along it.
For instance, some dependency paths including the one concerning
{\prg X=a} in \fig{pqprogtree} can be shortened by passing
predicate {\prg p} across clause (A).
To formulate this, let us introduce new predicate {\prg p1},
which is to be obtained by extracting {\prg p} from itself.
That is, we add the following entry to the memoization table:
\onecitm[p1]{\prg \NEG p1 $\equiv$ \SOME X\{\NEG p(X) \AND\ p(X)\}}
Now we pass {\prg p} through (A)
to new predicate {\prg q1}, replacing (A) with (A$'$):
\begin{itemz}
\item[(A$'$)] {\prg \LA\ p1 \AND\ q1(U,Y) \AND\ r(Y,Z).}
\end{itemz}
{\prg q1} has been dynamically defined here as below:
\onecitm[q1]{\prg q1(U,Y) $\equiv$ \SOME X\{q(U,X,Y) \AND\ p(X)\}}
This is entered to the memoization table, too.
Note that (A$'$) is equivalent to (A).

Next let us further pass {\prg p} across (D) and across (E),
and generate new clauses (D$'$) and (E$'$), respectively:
\begin{itemz}
\item[(D$'$)] {\prg q2(U) \LA\ U=b.}
\item[(E$'$)] {\prg q1(U,Z) \LA\ U=g(V) \AND\ q1(V,Y) \AND\ Y=f(Z).}
\end{itemz}
Folding based on \pref{q1} occurs twice when deriving (E$'$) from (E).
New predicate {\prg q2} has been defined by the below new entry
of the memoization table:
\begin{quote}\prg
\NEG q2(U) $\equiv$ \SOME X\{\NEG q1(U,X) \AND\ p(X)\}
\end{quote}
Note that (D) entails (D$'$) and (E) entails (E$'$), but not vice versa.
So (D) and (E) remain instead of being replaced by (D$'$) and (E$'$).

Note also that {\prg p} has been passed through (D)
not only downwards but also upwards.
By further passing {\prg p} through (A$'$),
an additional top clause (A$''$) is derived from (A$'$):
\begin{itemz}
\item[(A$''$)] {\prg \LA\ p1 \AND\ q2(U) \AND\ r1(Z).}
\end{itemz}
New predicate {\prg r1} is defined by:
\begin{quote}\prg
r1(Z) $\equiv$ \SOME Y\{r(Y,Z) \AND\ p(Y)\}
\end{quote}

If only downward movement were allowed, we would have to move
literal {\prg p(X)} downwards into {\prg q(U,X,Y)} or vice versa in (A)
at the beginning, in order to shorten the dependency paths running through
{\prg X}.
Since the resulting literal is equivalent to {\prg q1(U,Y)}.
This time we must move it downwards into {\prg r(Y,Z)} or vice versa,
in order to shorten the dependency paths running through {\prg Y}.
In either case, however, we must create a new binary predicate.

In contrast, the above upward passing creates unary predicates {\prg q2}
and {\prg r1}.
In general, passing a predicate across another
derives a predicate whose arity is the total arity of the former two
predicates minus the shared arguments
(because no variable appears more than twice in a clause).
So let us posit the below optimization strategy
(which is effective even if a variable can occur more than twice in a clause)
to suppress the arities of dynamically created predicates:
\onecitm[minarity]{Move the predicate with the smallest arity
of those on a dependency path.}
This calls for both downward and upward passing, as the above example shows.

\begin{figure*}[htbp]
\def\labelheight{18ex}
\centerline{\epsfbox{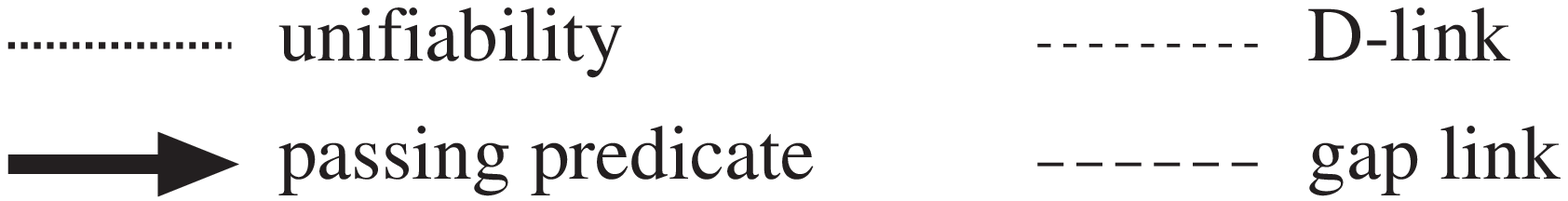}}
\epscap{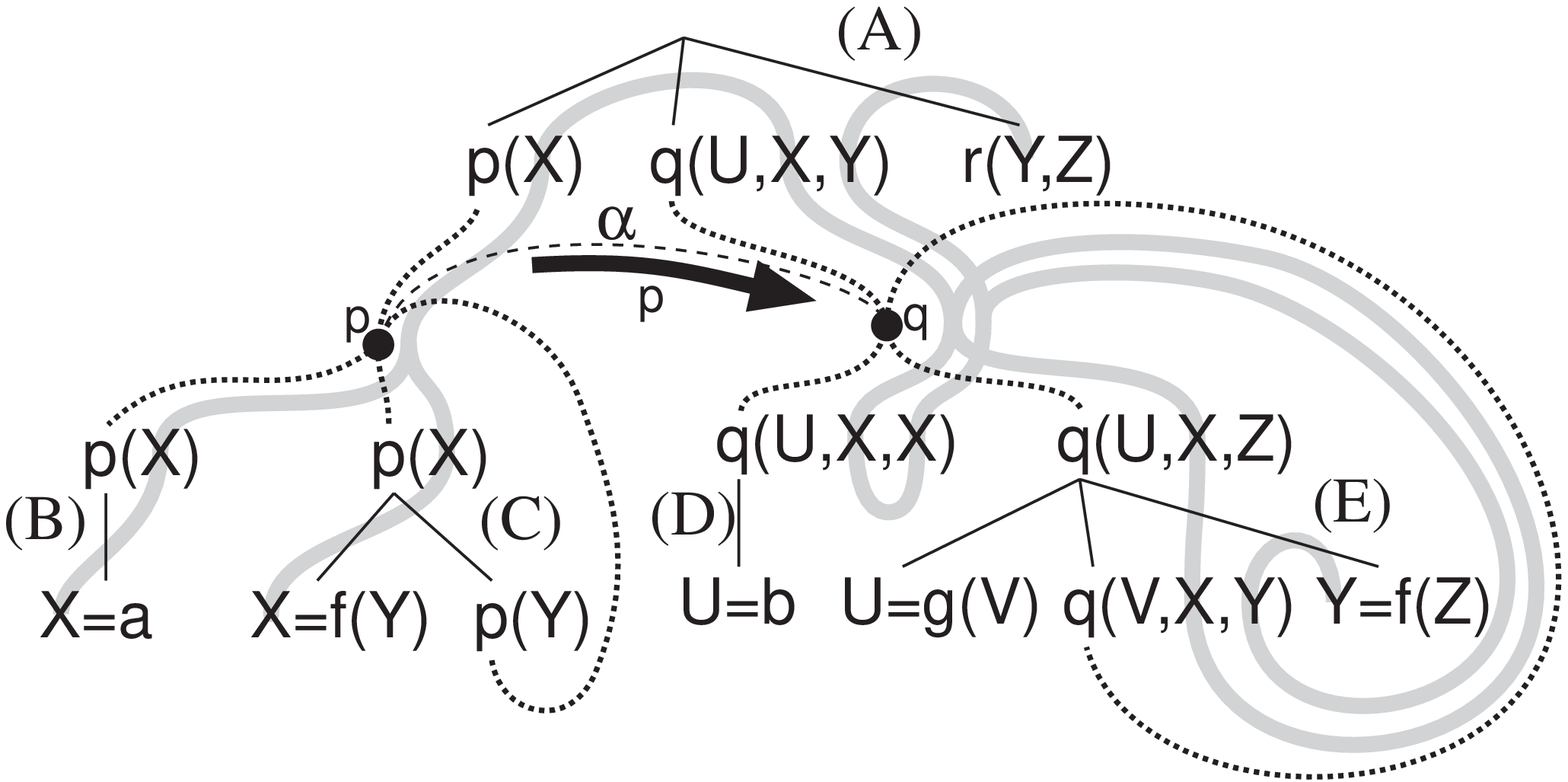}{i}
\epscap{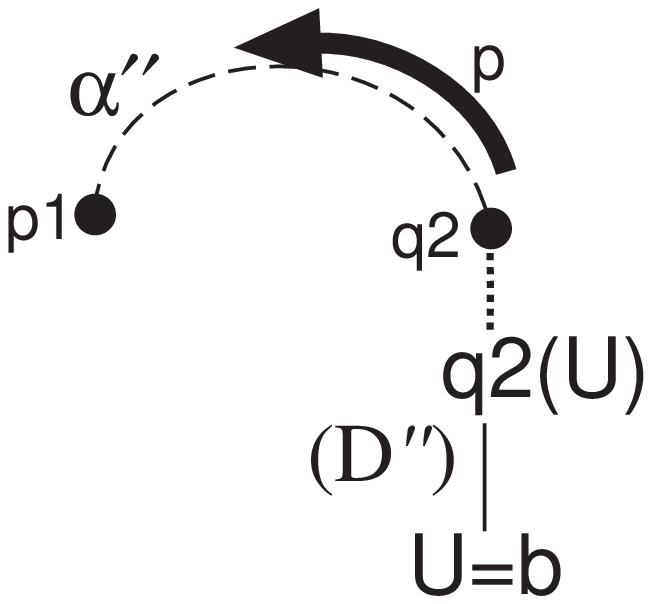}{ii}\epscap{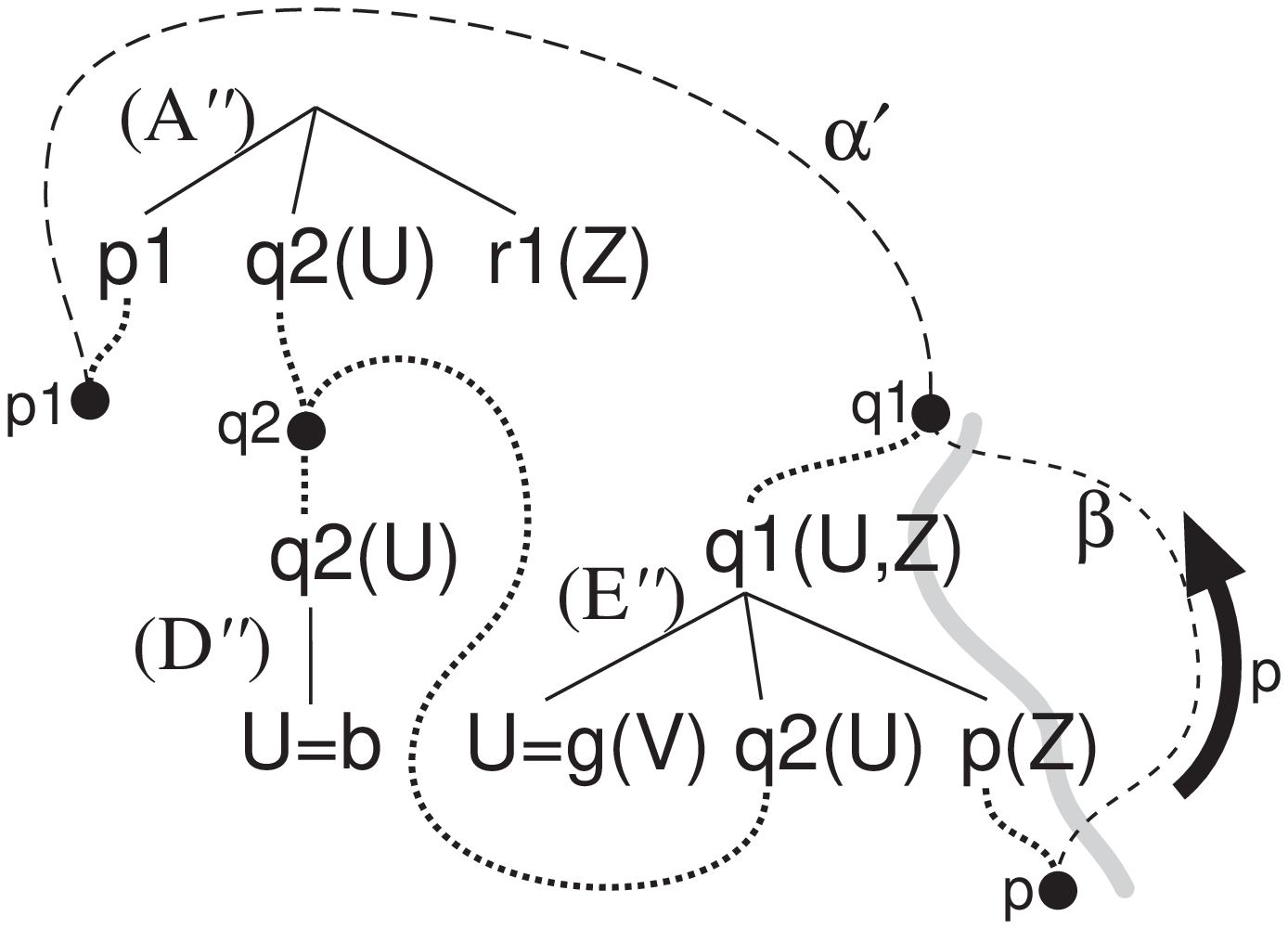}{iii}
\epscap{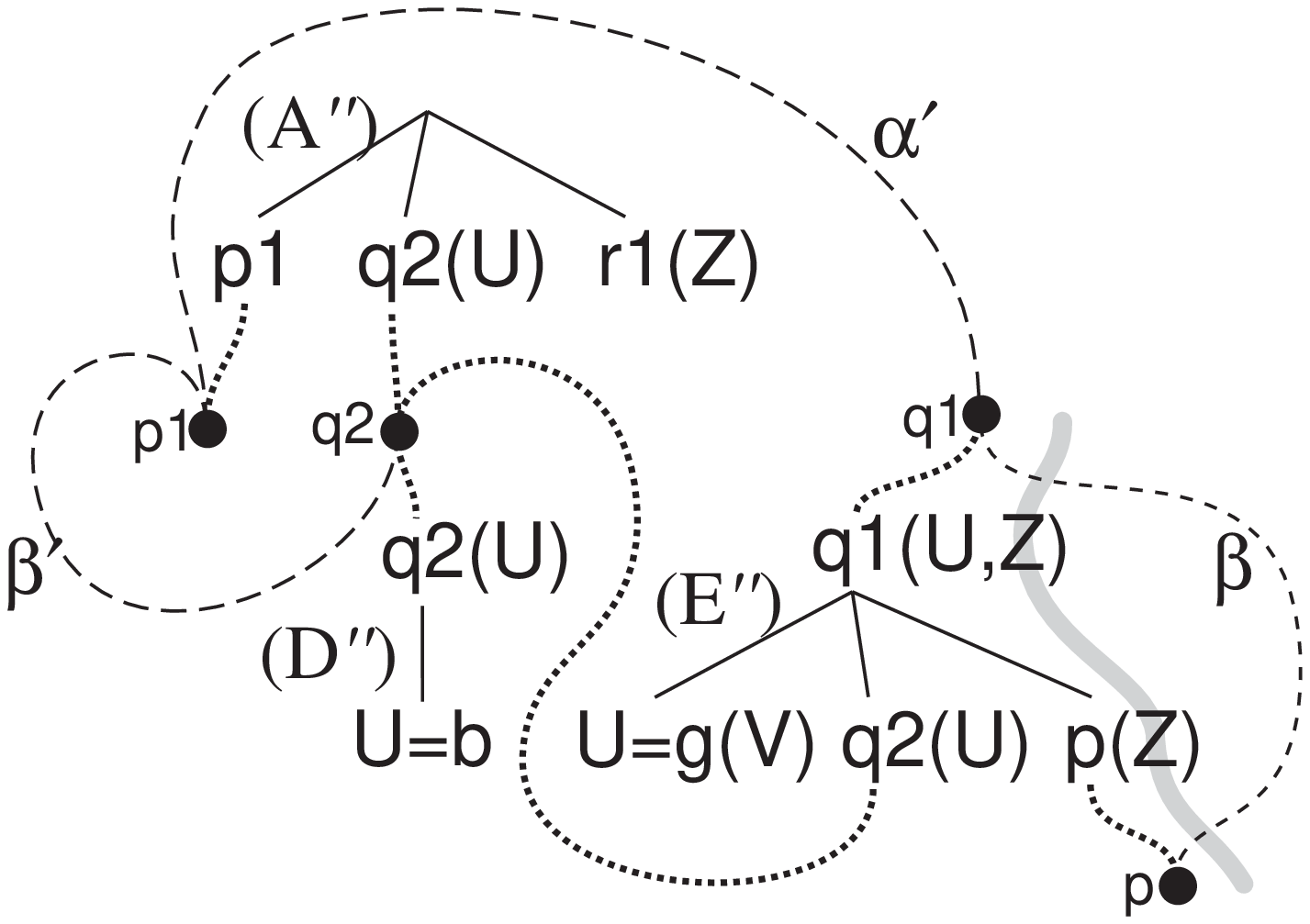}{iv}
\caption{Dependency reduction with compilation.}\label{pq}
\end{figure*}

Here let us turn to further optimization.
We do not always have to pass predicates through adjacent clauses,
but we can often pass them further at one stretch,
skipping the middle of dependency paths.
For instance, {\prg p} in \pref{p1} can be passed
into and across (D), skipping (A) and all the instances of (E).
This takes us from \fig{pq}%
(i) (the initial state of the program consisting of (A), (B), etc.) to (ii).
In \fig{pq}, a bullet ($\bullet$) connecting literals by
broken edges represents the predicate they share,
indicating that every upper (negative) literal
(a literal calling the predicate)
is unifiable with every lower (positive) literal
(the head of a definition clause).

The thin dotted edge $\alpha$ connecting {\prg p1} and {\prg q1} in (i)
is a {\bf dependency link}.
A dependency link connects two arguments of two predicates
and represents the set of dependency paths connecting those arguments.
Let us assume that $\alpha$ represents the set of
dependency paths connecting the argument of {\prg p}
and the second argument of {\prg q}.
These dependency paths go through (A) and instances of (E).
We can hence skip these dependency paths by passing {\prg p} along $\alpha$.

Such a skip gives rise to a gap in a program tree.
Each such gap is represented by a {\bf gap link}
which is derived from the dependency link
mediating the passing of the predicate.
$\alpha'$ in (ii) is a gap link
derived from $\alpha$ and represents the gap created
by the skip mentioned above.
Thus the gapped program tree (ii) represents all the program trees
obtained by filling the gap by (A$'$) and zero or more instances of (E$'$).

To be more precise,
we need more restricted notions of dependency path and dependency link,
together with some more relevant terms.
An {\bf active predicate} is the predicate to move.
An {\bf active literal} is a literal through which you can reach
an active predicate in the same (non-gapped) program tree.
For instance, {\prg q(U,X,X)} in clause (D) is an active literal
because you can go from (B) through {\prg q(U,X,X)} to {\prg p}.
On the other hand, {\prg \NEG q(V,X,Y)} (the body, negative, literal) is
inactive provided that predicate {\prg q} is inactive.
An {\bf active clause} is a clause containing more than two active literals.
There is no active clause in \fig{pq} (i)
if {\prg p} is the only active predicate.

We redefine a {\bf d-path} to be
a dependency path which connects two arguments
of two possibly different predicates, contains no U-turn,
and runs across no active predicate,
no active clause, and no two clauses with different numbers of active literals.
In \fig{pq} (i), for instance,
the path of arguments connecting the argument of {\prg p}
and the second argument of {\prg q} without going
through (D) is a d-path,
but that connecting the second and the third arguments of {\prg q} is not,
because it contains the U-turn in (D).
A {\bf D-path} is a non-empty set of d-paths
all running through the same sequence of clauses.
In \fig{pq}, there is no D-path consisting of
two or more non-null (that is, longer than zero) d-paths.
A {\bf D-link} is a link between two possibly different predicates
which represents the set of all D-paths which connect the same sets
of arguments of the two predicates.
We say a dependency path {\bf includes} another dependency path
when the former includes the latter as a subsequence.
We say a D-path {\bf includes} another D-path when every d-path
in the latter is included in some d-path in the former.
A D-path is {\bf maximal} if it is included in no other D-path.

As an invariant condition throughout the computation, we postulate:
\onecitm[dlink]{For every maximal D-path, there is a D-link containing it.}
Such a D-link is called a {\bf maximal D-link}.
In the current example, $\alpha$ is a maximal D-link.
Each maximal D-link is marked as such.
D-links are explicitly encoded as annotations to the program,
but D-paths are not.
We redefine here a gap link to be a derivative of a maximal D-link.
At one step of computation, a predicate is passed along a maximal D-link,
this D-link and the clause at its end derive a gap link and another clause,
respectively, and the predicate is inserted into the next predicate.
In the transition from \fig{pq} (i) to (ii), for instance,
predicate {\prg p} is moved across $\alpha$,
$\alpha$ derives a gap link $\alpha'$, clause (D) derives (D$''$),
and {\prg p} is inserted into new predicate {\prg q2}.

\pref{dlink} is to guarantee that maximal D-links exhaust the destinations
of the literals to move.
Some computations are necessary to maintain \pref{dlink},
as will be mentioned later.
The exclusion of U-turns from d-paths is to simplify
the representation of the program.
If U-turns were contained in D-links, apparently disjoint D-links may interact,
in the sense that a d-path included in one D-link
and one included in another may not coexist in a program tree,
which will complicate the representation of the program.
It is also for the sake of simplicity of representation that we
disallow a d-path to run across any active predicate,
any active clause,
and any two clauses with different numbers of active literals.
Namely, this is to guarantee that gaps are
appropriately represented by gap links.
Details are omitted because they are irrelevant as far as the examples
discussed in the present paper are concerned.

Back to the example, next we want to pass {\prg p} from {\prg q2}
as indicated by the thick arrow in \fig{pq} (ii).
However, this gapped program tree
does not contain any non-null D-path along which to move {\prg p} this way.
So we consult the part of the original program corresponding to the gap.
That is, we try to find where to move {\prg p} by looking at {\prg q},
because {\prg q} derived {\prg q2}.
Since $\alpha''$ has been derived from $\alpha$,
we must pass {\prg p} along some d-path included in $\alpha$;
otherwise the resulting gap cannot be represented by gap links.
We should hence pass {\prg p} to the end of maximal D-links
concerning the latter two arguments of {\prg q}.
Since the only such D-link is a null D-link,
we pass {\prg p} upwards with no skip.
So we check whether (A) and (E) are on any d-path represented by $\alpha$.
(A) meets this condition because {\prg X} is shared by
{\prg p(X)} and {\prg q(U,X,Y)} in it.
So does (E) because of its {\prg X}, too.
Thus (A$'$) replaces (A) and (E$''$) is derived from (E),
arriving at \fig{pq} (iii).
As mentioned before, it is because (A$'$) is equivalent to (A)
that (A$'$) replaces (A).
For simplicity, we have omitted the computation
to transform {\prg p(Y)} plus {\prg Y=f(Z)} to {\prg p(Z)},
but it is easy to formulate this as a primitive operation
because this computation concerns adjacent clauses only.
$\alpha'$ is a gap link derived from $\alpha$.

Note that we have made a new D-link $\beta$, which represent
the d-path connecting the argument of {\prg p} and the second
argument of {\prg q1}.
This is required by \pref{dlink},
because $\beta$ is a maximal D-link in this gapped program tree.
We have omitted another maximal D-link, which connects
the first argument of {\prg q1} and the argument of {\prg q2}.

From \fig{pq} (iii) we move {\prg p} along $\beta$, skipping (E$''$).
Then $\beta'$ is derived from $\beta$, as shown in (iv).
The computation is over here because there is no more dependency paths
connecting two non-variable terms.
The other definition clause of {\prg q2} can be derived from (E$''$) on demand,
though the current example involves no such demand.

D-links can be either statically precompiled or dynamically created.
In \fig{pq}, $\alpha$ has been precompiled
whereas $\beta$ is dynamically compiled.
D-links tend to raise the efficiency of the computation as a whole,
because one D-link usually represents several d-paths,
which can be skipped by following the D-link.
So let us employ this strategy:
\onecitm[compile]{Compile D-links and skip computation over them.}
D-links must be made to meet \pref{dlink} at least.
There are several alternatives about which non-maximal D-links to create.
Creating a D-link for every D-path is probably inefficient.
In what follows we assume that each D-link longer than one clause
is divided into two.
Namely, if a D-path $p$ running through two clauses
or more is represented by a D-link, then there are two D-links
representing two D-paths whose
concatenation is $p$ and which are at least one clause long.
Look at \fig{parse} (v) later for example.

\section{Left-Corner Parsing}

\begin{figure*}[htbp]
\def\labelheight{15ex}
\epscap{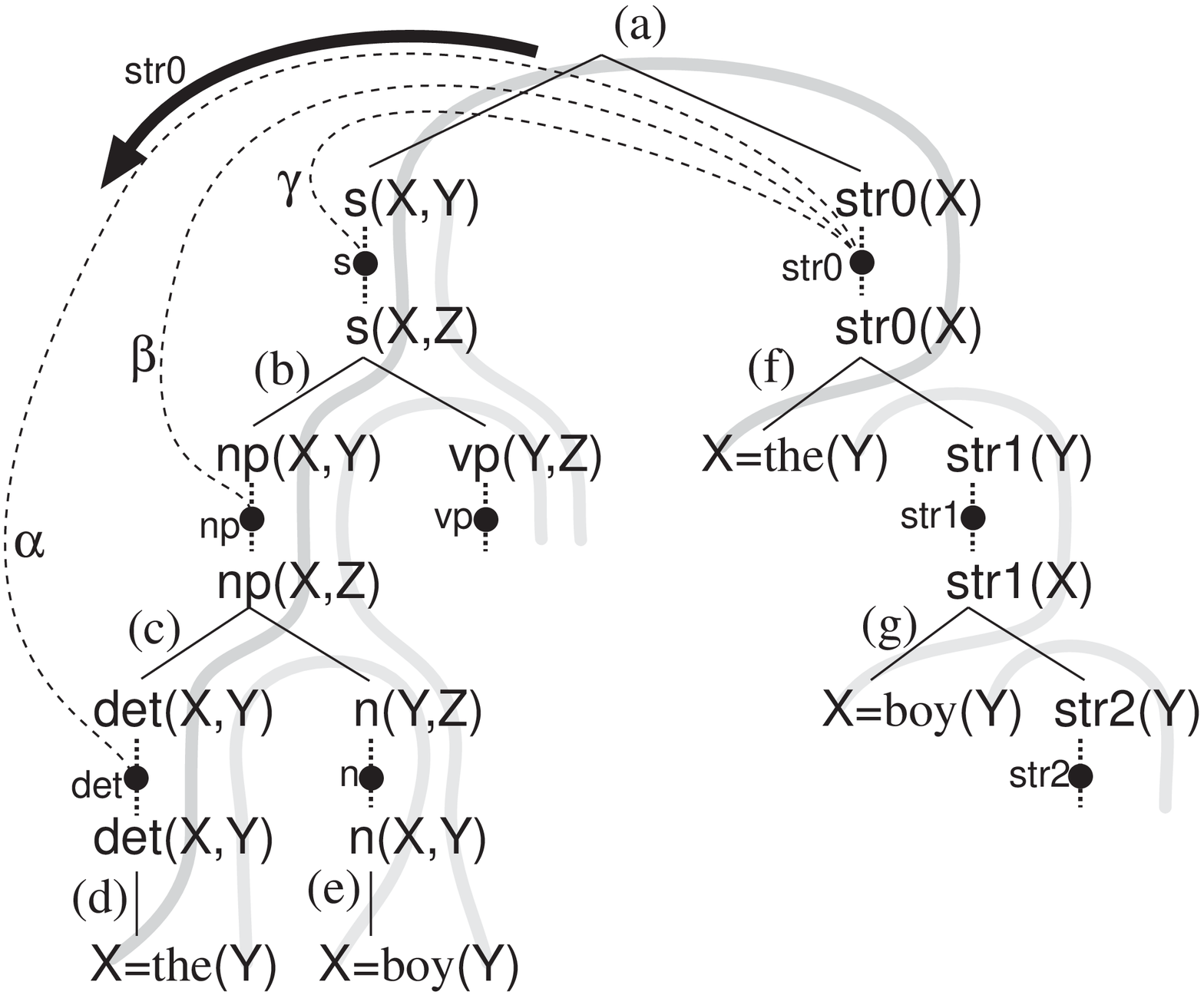}{i}\\
\epscap{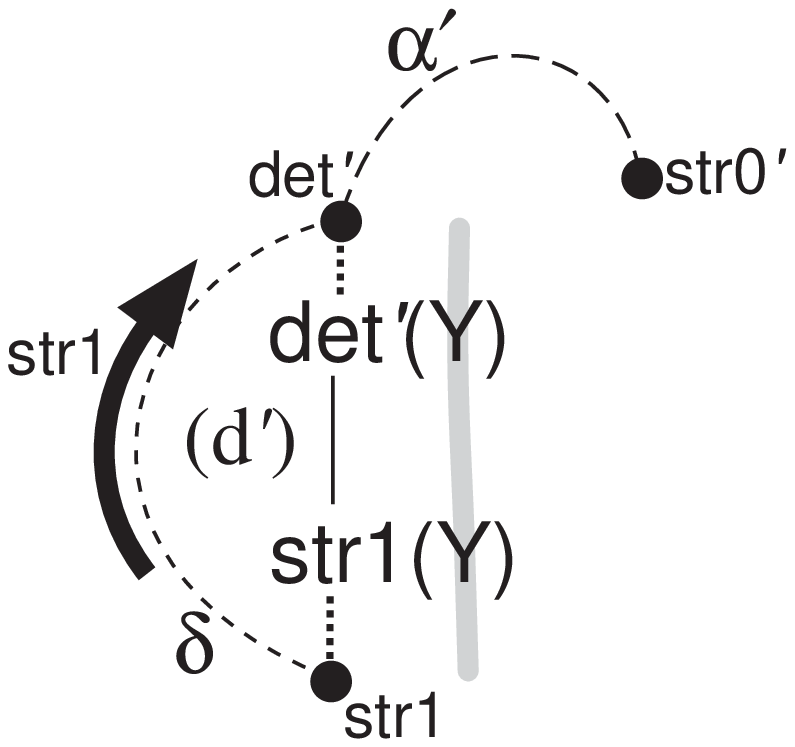}{ii} \epscap{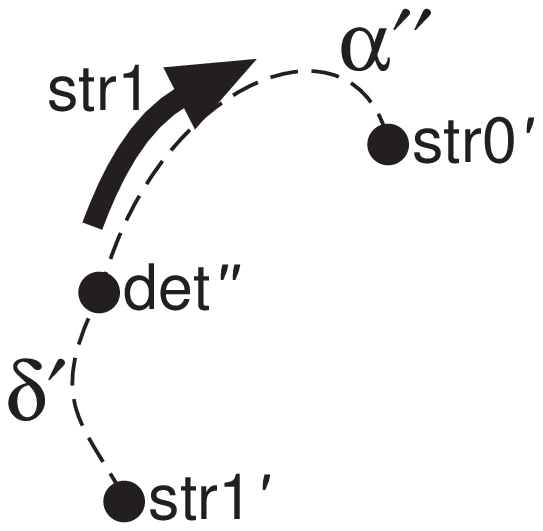}{iii} \epscap{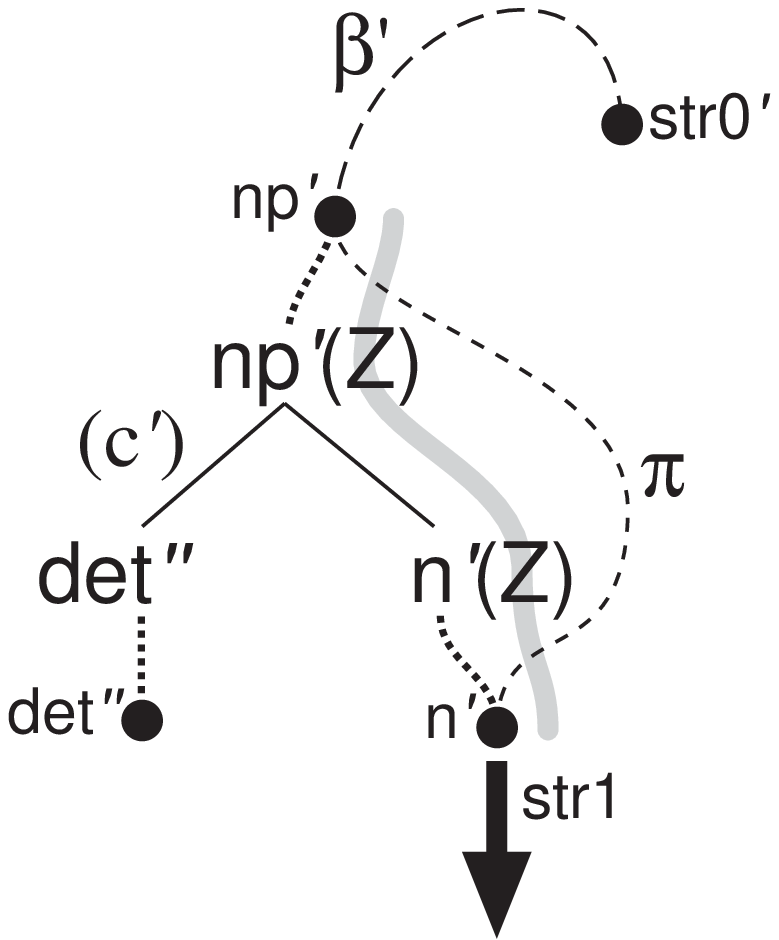}{iv}\\[2ex]
\epscap{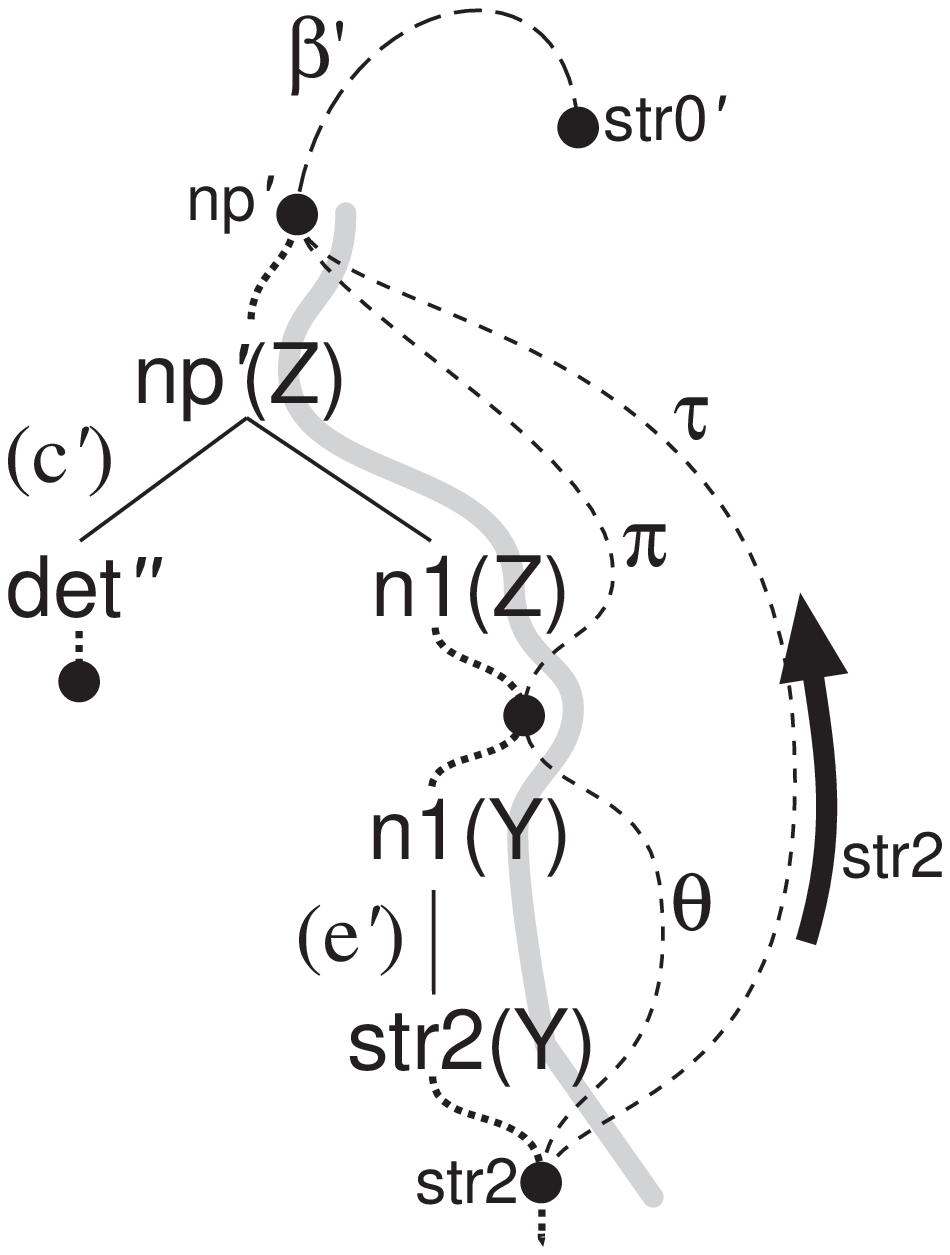}{v} \epscap{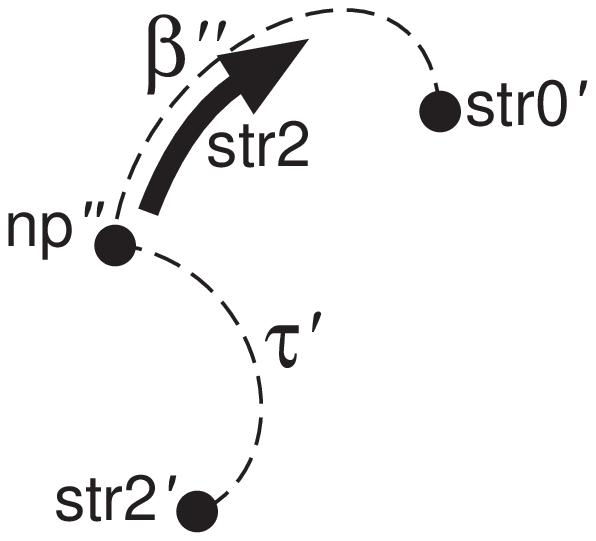}{vi} \epscap{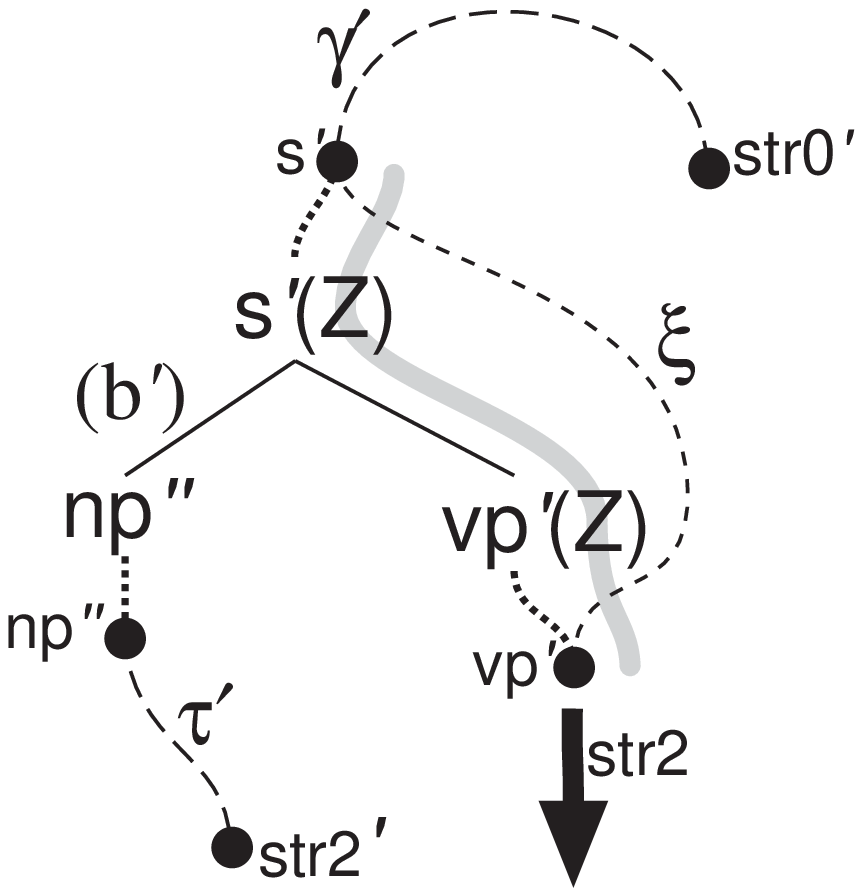}{vii}
\caption{Context-free parsing by dependency reduction.}\label{parse}
\end{figure*}

The following program
addresses parsing of a sentence beginning with `The boy.'
\begin{itemz}
\item[(a)]{\prg \LA\ s(X,Y) \AND\ str0(X).}
\item[(b)]{\prg s(X,Z) \LA\ np(X,Y) \AND\ vp(Y,Z).}
\item[(c)]{\prg np(X,Z) \LA\ det(X,Y) \AND\ n(Y,Z).}
\item[(d)]{\prg det(X,Y) \LA\ X=\auth{the}(Y).}
\item[(e)]{\prg n(X,Y) \LA\ X=\auth{boy}(Y).}\\
\CD
\item[(f)]{\prg str0(X) \LA\ X=\auth{the}(Y) \AND\ str1(Y).}
\item[(g)]{\prg str1(X) \LA\ X=\auth{boy}(Y) \AND\ str2(Y).}\\
\CD
\end{itemz}
This is depicted by \fig{parse} (i).
The left part ((a) through (e)) of the program tree
encodes the grammar and the right part ((f) and (g)) the input string.
In the following discussion and particularly at each entry in \fig{parse}
we consider just one (gapped) program tree for expository simplicity,
but of course there are lot more.
The structures under {\prg vp(Y,Z)} and {\prg str2(X)}
have been omitted for simplicity, too.

First we move {\prg str0} along D-link $\alpha$.
This derives clause (d$'$) from (d) and gap link
$\alpha'$ from $\alpha$,
reaching the gapped program tree in \fig{parse} (ii).
Thus the first dependency path connecting two non-variable terms,
has been eliminated and the second one has been recognized,
as indicated by the dim curve running through {\prg Y} of (d$'$).
This second dependency link is only partially instantiated,
but it may connect two non-variable terms because
{\prg det$'$(Y)} can be an active literal as {\prg det(X,Y)} is.
Also new maximal D-link $\delta$ representing this d-path across (d$'$)
has been created in order to meet \pref{dlink}.
New predicate {\prg det$'$} has been defined by the following entry
of the memoization table.
\begin{quote}\prg
det$'$(Y) $\equiv$ \SOME X\{det(X,Y) \AND\ str0(X)\}
\end{quote}

Next we go from (ii) to (iii) by moving {\prg str1} along $\delta$.
New predicate {\prg det$''$} is defined as follows:
\begin{quote}\prg
\NEG det$''$ $\equiv$ \SOME Y\{\NEG det$'$(Y) \AND\ str1(Y)\}
\end{quote}

From (iii) we want to pass {\prg str1} from {\prg det$''$} along $\alpha'$,
but $\alpha'$ does not subsume any d-path along which to pass {\prg str1}.
So the situation is similar to the one at \fig{pq} (ii).
We thus consult the part of the original program corresponding to the gap.
That is, we try to find where to move {\prg str1} by looking at {\prg det},
which derived {\prg det$''$}.
Since $\alpha''$ is a derivative of $\alpha$,
we must pass {\prg str1} along some d-path included in $\alpha$.
We should hence pass {\prg str1} to the end of maximal D-links
concerning the two arguments of {\prg det}.
Like in \fig{pq} (ii) again, the only such D-link is a null D-link.
Note that Clause (c) is on a d-path included in $\alpha$,
which is detected by noting D-link $\beta$ connecting
{\prg np} and {\prg str0}.
So we pass {\prg str1} through (c) to derive (c$'$),
as shown in \fig{parse} (iv).
Here we have created a maximal D-link $\pi$ by dynamic compilation.

From (iv) we can pass {\prg str1} downwards along any direction.
Passing it through clause (e), we eliminate the second dependency path
and thereby recognize the third, as in \fig{parse} (v).
$\theta$ and $\tau$ are D-links created now.
$\tau$ is the maximal D-link composed of $\pi$ and $\theta$.
{\prg str2} is hence moved along $\tau$, resulting in the gapped program
tree in (vi).
The transition from (vi) to (vii) is similar to that from (iii) to (iv).
The rest of the computation goes the same way.

Note that D-links serve both accessibility check and last-call optimization.
In the transition from (i) to (ii), that from (iii) to (iv),
and that from (v) to (vi), D-links serve as accessibility links.
An accessibility link connects a nonterminal symbol with another
which can appear as one of its leftmost descendant,
whereas a D-link connects two arguments (of predicates)
connected via one or more d-paths.
In the transition from (ii) to (iii) and that from (v) to (vi),
D-links allow last-call optimization,
which is often employed in programming language interpreters and compilers.

So this computation as a whole is essentially the standard left-corner parsing
with accessibility linking and last-call optimization, as shown in \fig{lcpar}.
\epsfig{lcpar}{Left-corner parsing.}
Each bullet is a word.
Each triangle is the Horn clause encoding a binary context-free rule.
The three edges of a triangle corresponds to the three variables
in the clause encoding points in the word string.
Broken lines mean that the rule is not yet applied.
Applied rules are drawn by solid lines,
and the eliminated variables
(variables through which predicates have been passed) are thick lines.
Due to precompiled D-links, a binary rule is not instantiated before
two variables in it are instantiated.
Due to dynamically created D-links, the rightmost variable
of each rule is not instantiated.
\fig{LCP1}
\epsfig{LCP1}{Schematic snapshot during parsing.}%
shows the gapped program tree in each moment of computation.

We have thus shown that
the two optimizing control strategies \pref{minarity} and \pref{compile}
derive the standard left-corner parsing.
In particular, precompiled dependency links serve as accessibility links,
and dynamically compiled dependency links support last-call optimization.
Left-corner parsing is hence derived from a general procedure
for constraint satisfaction, because the two strategies
are just for the sake of efficient elimination of dependency paths,
with no particular concern about any specific task (such as
parsing and generation) or domain (such as language).

\section{Other Examples}

Fixed computation order may be inefficient
even when the information source is homogeneous.
For instance, let us consider parsing based on a TAG (tree-adjoining grammar)
in a straightforward Horn-clause encoding such as below:
\onecitm[tag]{\prg a(X,Z,U,W) \LA\ b(X,Y,V,W) \AND\ c(Y,Z,U,V).}
{\prg a(X,Z,U,W)} represents an auxiliary tree
dominating two strings (differential lists)
{\prg X}-{\prg Z} and {\prg U}-{\prg W}.
This clause decomposes it into two smaller auxiliary trees
{\prg b(X,Y,V,W)} and {\prg c(Y,Z,U,V)}, as shown in \fig{TAG}.
\setbox\bx\hbox{\pref{tag}}%
\epsfig{TAG}{Graphical account of clause \usebox\bx.}%
In top-down parsing, {\prg a($a$,Z,U,W)} will be called where
$a$ is some postfix of the input string.
In OLDT, a problem arises here because
there are too many results of executing this literal:
The possible instantiations of {\prg Z} are not more than
the length of string {\prg X}, but there may be exponentially
or infinitely many values for {\prg U}-{\prg W}.
To avoid such combinatorial explosion,
we should pack these values incrementally,
which is essentially program transformation.

Dependency reduction does such incremental packing.
\pref{tag} is instantiated in two steps:
first by eliminating {\prg X} and {\prg Y} just in the case of
context-free parsing, and second by eliminating {\prg Z},
{\prg U}, and {\prg V}.
The space complexity of this parsing is hence $O(n^5)$,
$n$ being the sentence length, but it can be reduced to $O(n^4)$
by structure sharing among clauses \cite{epg}.
The time complexity is $O(n^6)$ because there are $O(n)$ ways of
making each final instance of the clause.

The two strategies \pref{minarity} and \pref{compile} also derive
semantic-head-driven generation \cite{shieber&90}.
In particular, \pref{compile} accounts for
the retrieval of a semantic head.
The accessibility link in head-corner parsing \cite{vannoord97}
is a special usage of D-link, too.

\section{Concluding Remarks}

We have proposed a constraint solver for Horn-clause programs,
called dependency reduction, and shown that it derives
standard efficient processes for parsing and generation.
A formal proof of completeness and soundness
of dependency reduction does not fit into the currently allocated space,
but we will report on it in an earliest possible opportunity.

Dependency reduction
can be straightforwardly extended to incorporate probability.
The key issue is attaching probability scores to D-links,
which employs a similar technique to \citeA{stolcke95}.
We would like to study a general method for controlling symbolic computation,
and natural language processing in probabilistic dependency reduction
will serve as an appropriate testbed for that.

\bibliographystyle{theapa}
\citepunct{}{\&}{and}{, }{; }{, }{}{}{.}
\bibliography{roman}
\end{document}